\documentclass[10pt, a4paper]{article}
\usepackage{lrec}
\usepackage{tipa}
\usepackage{graphicx}
\usepackage{tabularx}
\usepackage{soul}
\usepackage[official]{eurosym}
\usepackage{todonotes}
\usepackage{epstopdf}
\usepackage[latin1]{inputenc}

\usepackage{hyperref}
\usepackage{xstring}

\title{ \textbf{MASRI-HEADSET: A Maltese Corpus for Speech Recognition}}

\name{Carlos Mena, Albert Gatt, Andrea DeMarco, Claudia Borg, Lonneke van der Plas, \\ \textbf{\large{Amanda Muscat, Ian Padovani}}}

\address{University of Malta \\
         \{carlos.hernandez, albert.gatt, andrea.demarco, claudia.borg, lonneke.vanderplas \\
         amanda.muscat.11, ian.padovani.16\}@um.edu.mt\\}

\abstract{
Maltese, the national language of Malta, is spoken by approximately 500,000 people. Speech processing for Maltese is still in its early stages of development. In this paper, we present the first spoken Maltese corpus designed purposely for Automatic Speech Recognition (ASR). The
MASRI-HEADSET corpus was developed by the MASRI project at the 
University of Malta. It consists of 8 hours of speech paired with text, recorded by using 
short text snippets in a laboratory environment. The speakers were
recruited from different geographical locations all over the Maltese
islands, and were roughly evenly distributed by gender. This paper
also presents some initial results achieved in baseline experiments for Maltese ASR using Sphinx and Kaldi. The MASRI-HEADSET Corpus is publicly available for research/academic purposes.\\ \newline
}

\begin{document}

\maketitleabstract
%%%%%%%%%%%%%%%%%%%%%%%%%%%%%%%%%%%%%%%%%%%%%%%%%%%%%%%%%%%%%%%%%%%%%%%%%%
\section{Introduction}\label{sec:intro}
As digital resources and tools for Natural Language Processing and language-enabled interfaces become ever more common in daily use and commercial applications, it has become increasingly important to ensure that all languages are adequately represented in the digital sphere. This concern is evident, for example, in a recent resolution passed by the European Parliament to safeguard {\em language equality} in the digital age \cite{ep2018}.
 
In this regard, so-called `under-resourced' or `low-resourced' 
languages have been an ongoing concern within some sectors of the NLP 
community. For example, in a series of white papers published in
2011-12, reviewing the level of digital support for 31 of the 
languages of the European Union, 
the {\sc metanet} 
Initiative\footnote{\url{http://www.meta-net.eu/}} concluded that 
for a significant number of these languages, support in most areas 
of NLP was at best fragmentary and in some cases, weak or non-existent. 

The {\sc metanet} papers took the step of determining the level of support for different languages with reference to specific tasks or domains, namely: machine translation, text analysis, speech/language resources and speech processing. In principle, a language could have strong support in a subset of these, with weak or fragmentary support in others. This is reminiscent of the strategy used by \newcite{krauwer2003basic}, who discussed the notion of an `under-resourced' language in terms of a minimal set of language resources (a Basic Language Resource Kit, or {\sc blark}) necessary to undertake further precompetitive research and education. A somewhat more wide-ranging definition was more recently offered by \newcite{besacier2014automatic}, in their review of ASR for low-resourced languages, where the criteria included the following:

\begin{itemize}
    \item{Lack of a unique writing system or stable orthography.}
    \item{Lack of linguistic expertise.}
    \item{Limited presence on the web.}
    \item{Lack of electronic resources for speech and language processing.}
\end{itemize}

\subsection{The case of Maltese}
Maltese, the national language of Malta, was one of the languages which, at the time of the {\sc metanet} white papers, was ranked as having weak or no support under all four of the headings listed above \cite{Rosner2012}. This is in spite of the fact that Maltese does not suffer from the first two of the list of criteria offered by \newcite{besacier2014automatic}: the language not only has a long written tradition and a stable orthography \cite{azzopardi2013maltese}, but is very well-studied linguistically at all levels, including the morphological \cite[{\em inter alia}]{mifsud1995,hoberman2007maltese,gatt2018productivity}, syntactic \cite[{\em inter alia}]{Fabri1993,bulbul2018}, and phonological \cite{vella1994}, as well as in terms of its historical development \cite{brincat2011} and typological status \cite{comrie2009maltese}. On the other hand, its web presence, while comparatively small compared to that of languages such as English, could be argued to be proportional to the size of its community of speakers.

It is the fourth criterion listed by \newcite{besacier2014automatic}  -- lack of electronic resources -- that was the basis for the conclusions reached by \newcite{Rosner2012}. There are numerous factors which could have contributed to this situation. First, Maltese is a language with a small number of speakers (ca. 500,000); this fact makes NLP for Maltese appear less economically advantageous, at least for commercial developers. Second, Malta is officially a bilingual country, with English as the second language. The lack of digital support for Maltese has meant that many users resort to English for their electronic and online communication.  

On the other hand, this also implies that people whose language proficiency is Maltese-dominant might simply abstain from communicating through certain 
channels where their native language is not represented. Indeed, where speech is concerned, this is arguably also true for English speakers in Malta, since available speech interfaces, for example on mobile devices, tend to be developed for the recognition or synthesis of varieties of English (such as British, American or Australian) with a larger number of speakers, while Maltese English, which has been argued to be a variety in its own right \cite{grech2014}, is unsupported.

As a matter of fact, digital support for the Maltese language has improved drastically since the publication of the {\sc metanet} white paper (and this is likely true for a number of other languages covered by the white paper series). 
To take some examples, large annotated corpora for Maltese are now available, with accompanying tools for segmentation and labelling, including tokenisation and part-of-speech annotation \cite{gatt2013digital}.\footnote{Many of these resources are available through the Maltese Language Resource Server: \url{https://mlrs.research.um.edu.mt}}. Advances have been made in the development of electronic lexicons \cite{camilleri2013} and in automatic morphological analysis and labelling \cite{borg2017morphological,acling2017} as well as dependency parsing \cite{tiedemann2016,zammit2018}. 

Advances in speech technology for Maltese have however been comparatively limited. While there have been successful attempts to build speech synthesis systems using concatenative techniques \cite{micallef1997text,borg2011preparation}, no tools currently exist for Automatic Speech Recognition (ASR). This is partly due to a substantial data bottleneck where resources for speech engineering are concerned. 

\subsection{Aims of the present paper}
The present paper addresses this gap, presenting a new corpus for ASR, built in the context of ongoing work in the {\sc masri} (Maltese Automatic Speech Recognition) project. The paper describes the {\sc masri-headset} Corpus ({\sc mhc}), the first 
corpus in Maltese suitable for training ASR systems.

The rest of this paper is structured as follows. Section \ref{sec:corpus-desc} describes the design strategy and the recording process leading up to the corpus. In Section \ref{sec:experiments}, we demonstrate 
its suitability for creating acoustic models for ASR by running some experiments in CMU-Sphinx\footnote{An open source speech recognition toolkit. See 
\url{https://cmusphinx.github.io/}} and 
Kaldi\footnote{An open source speech recognition 
toolkit. See \url{https://kaldi-asr.org/index.html}}. We also discuss the construction of pronunciation models of Maltese. Section \ref{sec:conclusion} concludes the paper with a brief overview of ongoing work on expanding Maltese speech resources, and on deploying recent techniques for ASR that reduce the dependency on large data sources.

%%%%%%%%%%%%%%%%%%%%%%%%%%%%%%%%%%%%%%%%%%%%%%%%%%%%%%%%%%%%%%%%%%%%%%%%%%
\section{The MASRI-HEADSET Corpus (MHC)}\label{sec:corpus-desc}

MASRI-HEADSET is the first Maltese corpus specifically designed with ASR systems in mind. While it is comparatively small, it constitutes the first step in the creation of larger-scale resources for speech recognition in Maltese.

In this section, we describe the design of MHC, including the selection of participants whose voices 
were recorded for the corpus, as well as the process whereby textual 
prompts were selected. Finally, we discuss the characteristics
of the final release version of the corpus.

%------------------------------------------------------------------------%
\subsection{Corpus Design and Collection}

\begin{figure}
    \centering
    \includegraphics[scale=0.3]{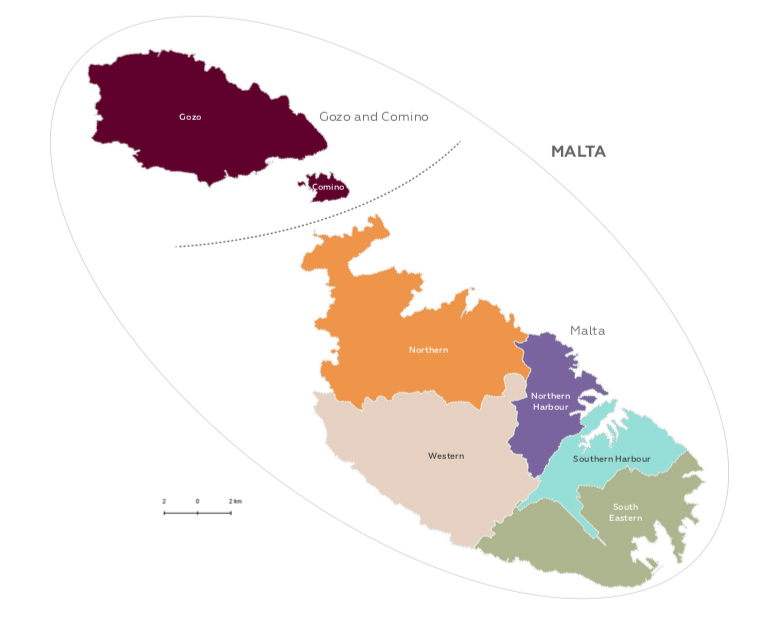}
    \caption{Regions in the Maltese islands \protect\cite{nso2019}}
    \label{fig:regions}
\end{figure}

\paragraph{Participants} Expressions of interest in participation were 
solicited via online advertisements. Initially, 61 people 
expressed interest. In an effort to ensure adequate representation of 
accent and speaker variation, participants were selected to achieve a 
balanced sample in terms of gender, age and geographical location, the 
latter being defined in terms of the six regions of the Maltese islands 
defined by the National Statistics Office for data collection 
purposes \cite{nso2019}, as shown in Figure \ref{fig:regions}. Out of a 
total of 61 people who expressed interest, 25 individuals 
were recruited. The sample is approximately gender-balanced, with 13 female and 12 male speakers (age range = 18 to 31; mean age = $23.9$, SD = $3.54$). Table \ref{tbl:overview} provides a summary of the participants and the average number of hours and utterances split by gender. Participation 
was remunerated at the rate of \euro{15} for approximately one hour 
of recording. Written consent was obtained and the procedure was 
screened by the procedures of the University of Malta Research Ethics Committee.\footnote{\url{https://www.um.edu.mt/urec}}

\begin{table}[ht]
\begin{center}
\begin{tabular}{|l|l|l|}\hline
                   & {\bf Female} & {\bf Male} \\
\hline                   
Number of Speakers & 13 & 12  \\
Average amount per speaker & 19m:57s & 18m:53s \\
Average utterances per speaker & $158.0$ & $150.8$ \\\hline
\end{tabular}
 \caption{Overview of the MASRI-HEADSET Corpus}
 \label{tbl:overview}
\end{center}
\end{table}

\paragraph{Language dominance:} Given the bilingual situation in Malta, 
where apart from Maltese, English is also an official language, and where 
some individuals are English-dominant, prospective participants were 
asked to respond to an online language background questionnaire. The 
25 participants selected were all Maltese-dominant speakers, with 
some also speaking dialects of Maltese, in addition to standard 
Maltese. Speakers with dominance in English rather than Maltese were not 
eligible for participation in this particular corpus collection.

\paragraph{Sentence selection:} Approximately 200 samples of sentential prompts were selected from the 
Korpus Malti v3.0\footnote{\url{http://mlrs.research.um.edu.mt/}} \cite{gatt2013digital}. Samples were built using the 
following procedure, which was intended to ensure that each 
sample would contain as broad a range of phonological and phonotactic 
variation as possible:

\begin{enumerate}
    \item The full corpus was transcribed automatically using a rule-based
    grapheme-to-phoneme (G2P) mapping procedure;
    \item A trigram model over phone sequences was constructed based 
    on the full corpus;
    \item The corpus text was broken up into blocks of approximately 75 
    orthographic words each, with each block paired with its orthographic 
    transcription. Blocks were randomly shuffled. Samples were then constructed by greedily adding blocks to a sample, as per the following algorithm:
    \item For $i = 1$ to $n$ samples, with size $m$:
    \begin{enumerate}
        \item initialise sample $s_i$ to $\emptyset$
        \item While $\vert s_i \vert < m$, sample the next best block 
        and include it in $s_i$
    \end{enumerate}
\end{enumerate}

where the `next best block' is defined in terms of coverage: given 
the current distribution of trigrams in blocks already included in 
the sample, the best next block is the one that results in the 
greatest amount of variation in the sample once it is included. Once a sample was constructed, prompts containing proper names and/or words in English were pruned.

The result of this procedure was a set of samples, each consisting 
of blocks of text (our prompts). Each of the 25 participants was 
required to read each block of text in the sample assigned to them.

\paragraph{Recording methods: }
Speech recordings were made in a quiet room at the Faculty of ICT 
at the University of Malta. The recordings were done in a dual 
fashion: close (using a headset) and far-field (via a microphone 
at a distance of approximately 3 metres from the participant). Only 
the headset data is included in the release under discussion. 

Each speaker read prompts from one of the samples, shown on a 
computer screen in random order using SpeechRecorder \cite{draxler2007speech}, which was also used to capture the recordings. The headset recordings were captured using a Sennheiser PC-8 (48KHz) mic/headset and the recording sessions typically lasted between 45 minutes and one hour for each of the participants.

\subsection{Characteristics of the Release Version}
\label{sec:CorpusChars}
After recording, the data underwent a process of pruning and normalisation, resulting in the following characteristics:

\begin{itemize}
    % \item{In the selection process of the utterances for the release 
    % version of the MHC, utterances with words in English and with 
    % proper-names were avoided.}
    \item{The MHC has an exact duration of 8 hours and 6 minutes. It has 
    3864 audio files.}
    \item{Every audio file contains only the voice of one single 
    speaker with no background noise.}
    \item{Utterances with stuttering and/or mispronunciations were 
    pruned.}
    % \item{The MHC has recordings from 25 different speakers: 13 women 
    % and 12 men.}
    \item{Data in MHC is classified by speaker, with all 
    recordings of one speaker stored in a separate directory.}
    \item{Data is also classified according to the gender (male / female) 
    of the speakers.}
    \item{Audio files in the MHC are distributed in a 16khz @ 16bit mono 
    format.}
    \item{The MHC corpus contains 3864 utterances (one per audio file), with a total of 11,503 unique words or tokens.}
    \item{Every audio file has a unique ID that is compatible with ASR engines such as Kaldi~\cite{povey2011kaldi} and  CMU-Sphinx~\cite{lamere2003cmu}.}
    \item{All textual transcriptions in MHC are lowercased. All punctuation is removed, with the exception of hyphens (-) and apostrophes ('), which are part of Maltese orthography.}
\end{itemize}

Table \ref{tbl:detail} provides a detailed overview of each participant in terms of the total time recorded and the number of utterances read. 

\begin{table}[ht]
\begin{center}
\begin{tabular}{|l|l|c|}\hline
Speaker ID & Total time & No. of Utterances\\
\hline
    F\_01	& 12m:57s &	112 \\
    F\_02	& 13m:27s &	127 \\
    F\_03	& 24m:19s &	182 \\
    F\_04	& 20m:48s &	166 \\
    F\_05	& 19m:35s &	174 \\
    F\_06	& 18m:11s &	125 \\
    F\_07	& 22m:56s &	161 \\
    F\_08	& 19m:29s &	171 \\
    F\_09	& 21m:15s &	170 \\
    F\_10	& 23m:16s &	169 \\
    F\_11	& 17m:46s &	144 \\
    F\_12	& 21m:18s &	178 \\
    F\_13	& 24m:10s &	175 \\
    
    M\_01	& 15m:20s &	118 \\
    M\_02	& 22m:02s &	164 \\
    M\_03	& 15m:18s &	117 \\
    M\_04	& 17m:19s &	125 \\
    M\_05	& 22m:26s &	159 \\
    M\_06	& 17m:57s &	179 \\
    M\_07	& 21m:26s &	177 \\
    M\_08	& 19m:17s &	170 \\
    M\_09	& 20m:05s &	166 \\
    M\_10	& 18m:02s &	134 \\
    M\_11	& 20m:08s &	164 \\
    M\_12	& 17m:18s &	137 \\
\hline
\end{tabular}
 \caption{Details for MASRI-HEADSET Corpus}
 \label{tbl:detail}
\end{center}
\end{table}

%%%%%%%%%%%%%%%%%%%%%%%%%%%%%%%%%%%%%%%%%%%%%%%%%%%%%%%%%%%%%%%%%%%%%%%%%%

\section{Experiments}\label{sec:experiments}

In this section, we describe a number of experiments in constructing ASR systems using a typical architecture consisting of  language model, pronunciation model
and acoustic model. These are intended to provide baseline results, demonstrating that the MHC is suitable for training ASR systems using off-the-shelf toolkits, namely, the 
Pocketsphinx\footnote{Pocketsphinx is a real-time version of the CMU-Sphinx~\cite{lamere2003cmu}.}~\cite{huggins2006pocketsphinx} 
and
Kaldi~\cite{povey2011kaldi} engines. The choice of these toolkits was based on the following rationale. 

MHC is a relatively small dataset compared, for example, to standard datasets for English \cite{panayotov2015librispeech,Roter2019}. State of the art systems, such as DeepSpeech \cite{amodei2016deep}, Wav2Letter \cite{collobert2016wav2letter} or Espresso \cite{wang2019espresso} are known to require datasets which are orders of magnitude larger than MHC. At the same time, Sphinx and Kaldi remain widely used for quick prototyping, and perform a series of transformations on input data that make them less data hungry than more sophisticated systems. Since our aim is to produce an initial set of baseline results, these toolkits were appropriate.

%(CB: I don't think we need to say this right now) 
%At the end of the paper, in the appendix, we will instructions on where to download the files needed to reproduce these experiments.

%------------------------------------------------------------------------%
\subsection{Training and test data}
For the purposes of these experiments, the MHC was randomly divided into training (3614 files, totalling 7h:35m:21s) and test (250 files, totalling 30m:57s) sets. Note that the corpus, as distributed, does not reproduce the split, but is distributed as one whole set. Nevertheless, one can reproduce the experiments of this paper with
the help of our "LREC2020 Experiment Files" available in our
project website\footnote{\url{https://www.um.edu.mt/projects/masri/downloads.html}}.

While the selected test set is small, it was sampled to ensure representativeness, by including 10 audio files from each of the 25 speakers represented in the corpus. The aim is thus to enable an evaluation of a relatively simple ASR model on test data with limited speaker variation (in the sense that data from all speakers is also represented in the training set).

In constructing a language model and pronunciation dictionary for the experiments reported below, none of the utterances or lexical items in the test set were included.

\subsection{The Language Model}
\label{sec:LangModel}
The language model was created using part of the 
Korpus Malti v3.0 \cite{gatt2013digital}, a corpus of written 
or transcribed Maltese divided into different genres, 
including: culture, news, academic, religion, sports, 
etc. The corpus is annotated with part of speech information 
and has a size of approximately 250 million 
tokens. A substantial proportion of the tokens are 
also lemmatised. 

For the purposes of these experiments, sentences containing 
Maltese-English code-switching or extensive borrowing from
English were excluded -- Engish words were identified via 
the CMU Pronounciation 
Dictionary \cite{weide1998cmu}. We also 
removed sentences with 
digits and proper names (to distinguish 
between proper names and other tokens, we used CIEMPIESS-PNPD
\cite{mena2019opencor}). After the selection process, we ended 
up with more than 28,000 sentences. In addition, we included the transcriptions from the MHC training data (3,614 in total) so as to reduce perplexity. 

%the heuristic strategy also used to select utterances (see %Section~\ref{sec:CorpusChars}).
%Code switching between Maltese 
%and English is a very common phenomena in Malta and this is also reflected in the corpus. Nevertheless, for these 
%particular experiments we decided to discard (as much as possible) all  
% sentences with non-Maltese words, following the same strategy used to 
% select the utterances discussed in section~\ref{sec:CorpusChars}. 

Using the above data, a 3-gram language model was produced using the SRI Language Modelling Toolkit.\footnote{\url{http://www.speech.sri.com/projects/srilm/}}

For the experiments with Sphinx, the model could be directly included in ARPA format; for Kaldi, the format was transformed using the off-the-shelf Kaldi executable {\tt arpa2fst}.

%------------------------------------------------------------------------%
\subsection{The Pronunciation Model}

\begin{figure}[!ht]
    \begin{center}
        \includegraphics[scale=0.3]{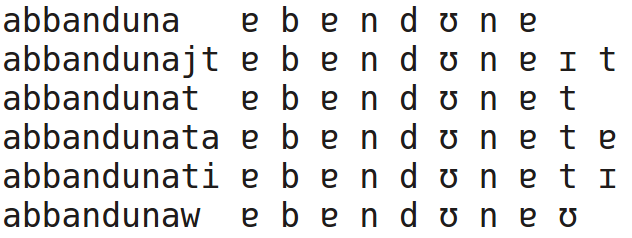}
        \caption{Format of the Pronounciation Dictionary}
        \label{fig:PronDict}
    \end{center}
\end{figure}

Traditional ASR systems, such as Sphinx, Kaldi or 
HTK \cite{young1993htk}, require a
pronunciation model in the form of a pronunciation dictionary. The
format of these kind of dictionaries is very simple: a list of 
words alphabetically sorted, each of them followed by a sequence
of phonemes separated by spaces. An example is shown in figure \ref{fig:PronDict} for the lemma {\em abbanduna} `to abandon' and its derivations.

Words in the Korpus Malti v3.0 corpus were converted into their phonetic transcription using a grapheme-to-phoneme tool (G2P), which is a new Python 3 implementation of an earlier model by \newcite{borg2011preparation}, originally built in the context of a text-to-speech system for Maltese (see Section \ref{sec:intro}). The phoneme set used by the G2P tool is given in the Appendix, Table~\ref{tbl:Maltese_Phonology_Table}. 
% Their G2P tool was released as a binary file that could only run on a Windows operating system, so we decided to re-implement the G2P tool using the same rules in Python 3. 
% The pronunciation of each word was based on the grapheme-to-phoneme conversion tool (G2P tool). Our G2P tool is based on grapheme-to-phoneme mapping rules developed by KPMG Microsoft Business Solutions (formerly CrimsonWing) in the context of an earlier project on concatenative text-to-speech synthesis for Maltese \cite{borg2011preparation}.

The resulting pronunciation dictionary contains more than 4,000 words. 
% The set of phonemes  
% the whole set of phonemes used by our G2P tool is specified in 
% Table. 
Table~\ref{tbl:PhonemeDistro} shows 
the phoneme distribution in the MHC corpus.

 \begin{table}
 \begin{center}
 \begin{tabular}{|c|c|c|c|}
    \hline
        No. &       Phoneme                 &   Counts  &   Percentage	\\
    \hline
		 1	&	\textturna	                &	24340	&	$12.1790$\%	\\
		 2	&	\textsci                	&	22622	&	$11.3193$\%	\\
		 3	&	l	                        &	14133	&	$7.0717$\%	\\
		 4	&	t                       	&	14087	&	$7.0487$\%	\\
		 5	&	n       	                &	11227	&	$5.6176$\%	\\
		 6	&	\textupsilon               	&	11155	&	$5.5816$\%	\\
		 7	&	\textepsilon            	&	10920	&	$5.4640$\%	\\
		 8	&	m	                        &	9286	&	$4.6464$\%	\\
		 9	&	r       	                &	8597	&	$4.3017$\%	\\
		10	&	k       	                &	7380	&	$3.6927$\%	\\
		11	&	s	                        &	6679	&	$3.3420$\%	\\
		12	&	\textopeno	                &	5420	&	$2.7120$\%	\\
		13	&	h       	                &	5383	&	$2.6935$\%	\\
		14	&	j       	                &	5285	&	$2.6444$\%	\\
		15	&	d                       	&	5281	&	$2.6424$\%	\\
		16	&	\textsci\textipa{:}     	&	4946	&	$2.4748$\%	\\
		17	&	b	                        &	4575	&	$2.2892$\%	\\
		18	&	f	                        &	4573	&	$2.2882$\%	\\
		19	&	\textglotstop	            &	3881	&	$1.9419$\%	\\
		20	&	\textturna\textipa{:}   	&	3800	&	$1.9014$\%	\\
		21	&	\textesh                	&	3309	&	$1.6557$\%	\\
		22	&	p	                        &	2877	&	$1.4396$\%	\\
		23	&	w	                        &	1953    &	$0.9772$\%	\\
		24	&	\textdyoghlig           	&	1820	&	$0.9107$\%	\\
		25	&	z	                        &	1304	&	$0.6525$\%	\\
		26	&	\textteshlig	            &	1151	&	$0.5759$\%	\\
		27	&	\textepsilon\textipa{:} 	&	870	    &	$0.4353$\%	\\
		28	&	g	                        &	769	    &	$0.3848$\%	\\
		29	&	i\textipa{:}                &	574 	&	$0.2872$\%	\\
		30	&	v                       	&	480	    &	$0.2402$\%	\\
		31	&	\texttslig	                &	425	    &	$0.2127$\%	\\
		32	&	\textopeno\textipa{:}	    &	359	    &	$0.1796$\%	\\
		33	&	\textdyoghlig               &	136	    &	$0.0681$\%	\\
		34	&	\textyogh                  	&	129     &	$0.0645$\%	\\
		35	&	\`a	                        &	92	    &	$0.0460$\%	\\
		36	&	\`e                        	&	21	    &	$0.0105$\%	\\
		37	&	\`i	                        &	7   	&	$0.0035$\%	\\
		38	&	\`u	                        &	6	    &	$0.0030$\%	\\
		39	&	\`o	                        &	1   	&	$0.0005$\%	\\
    \hline

    \hline
 \end{tabular}
 \caption{Phoneme Distribution of the MASRI-HEADSET}
 \label{tbl:PhonemeDistro}
 \end{center}
 \end{table}
 
 Table~\ref{tbl:PhonemeDistro} reveals that there is no point in 
 considering G2P rules for the low-pitch vowels due to their small 
 counts. While such vowels are known to help to improve naturalness in speech synthesis systems, their low frequency suggests they might have limited value for ASR. We leave this as a question for future work.
%  could be worthless in speech recognition if 
%  they are too few. We leave this discussion for future work and 
%  updates of the G2P tool.

%------------------------------------------------------------------------%
% \subsection{The Test Set}

% The MASRI-HEADSET is an small dataset compared to others that
% we can find at no cost on the web for English  
% \cite{panayotov2015librispeech,Roter2019}. For this reason, we 
% decided to perform our preliminary experiments with engines such 
% as Sphinx and Kaldi. Despite the rise of more sophisticated systems, 
% such as DeepSpeech \cite{amodei2016deep}, Wav2Letter \cite{collobert2016wav2letter} 
% or Espresso \cite{wang2019espresso}, these two are still popular in the 
% ASR community. They perform a series of transformations to the input 
% data in order to retrieve as much information as possible. This 
% paradigm makes these systems less ``data hungry'' than the more sophisticated 
% systems. 

% For the experiments, the selected test set is small but representative because it includes
% 10 audio files from all the 25 speakers of the whole corpus. The point 
% is to learn something about the behavior of the system with every
% speaker while not reducing the training set in a significant way.

%------------------------------------------------------------------------%
\subsection{Sphinx Setup}

For this experiment, Sphinx was configured using standard settings:
%in a traditional fashion: 
continuous Hidden Markov Models (HMM) with 3 states per HMM and 1,050 tied states. The number
of tied states was determined empirically by running the training
stage several times and selecting the model with the lowest Word Error Rate (WER).

As features, we computed MFCC vectors through Sphinx for all of the 13 coefficients, plus the first and second derivatives.

In addition to the above setup, we also performed additional 
%After getting a good performance with this setup, we also performed an
experiments applying Linear Discriminant Analysis (LDA) and Maximum Likelihood Linear Transform (MLLT).
%to improve results.

%------------------------------------------------------------------------%
\subsection{Kaldi Setup}

We carried out two experiments in Kaldi: one with traditional HMMs and the
other with neural networks using the \texttt{nnet4d3} recipe.\footnote{\texttt{nnet4d3} 
is an architecture composed of two p-norm layers with 1000 neurons in each
layer. The number of pdfs that correspond to the outputs of the neural 
network is 1504 \cite{lim2019integration}.} The following processes were used 
%processes implemented 
during HMM training:

\begin{itemize}
    \item{Linear Discriminant Analysis (LDA)}
    \item{Maximum Likelihood Linear Transform (MLLT)}
    \item{Speaker Adaptive Training (SAT)}
\end{itemize}

%The nnet4d3 consists of a neural network with 2 layers and 1000 neurons
%in each layer \cite{lim2019integration}. In the appendix, we share the 
%files to reproduce the experiments.

%------------------------------------------------------------------------%
\subsection{Results}

Table~\ref{tbl:Results} shows the results for all the experiments
performed, providing information about the Word Error Rate (WER)
and the Sentence Error Rate (SER). 
% It is important to 
% point out that none of the utterances in the test set were included 
% in the language model and none of its words were included in the 
% pronunciation dictionary. Therefore, 
% we can consider the test set as 
%  completely unseen to the training stage. 
The overall perplexity of the language model in all experiments was  
% perplexity\footnote{The language model was created and the perplexity 
% was measured using SRLIM available 
% at: \url{http://www.speech.sri.com/projects/srilm/}.} in all the 
% experiments is 
$182.3744$, with 87 out-of-vocabulary (OOV) words.

 \begin{table}[ht]
 \begin{center}
 \begin{tabular}{|l|c|c|}
    \hline
        Experiment           & WER       & SER        \\
    \hline
        Sphinx HMMs          & $19.40$\% & $64.00$\%  \\
        Sphinx HMMs+LDA/MLLT & $18.40$\% & $59.60$\%  \\
        Kaldi HMMs           & $12.54$\% & $39.60$\%  \\
        Kaldi NNs            & \textbf{$10.56$\%} & \textbf{$37.60$\%}  \\
    \hline

    \hline
 \end{tabular}
 \caption{Results of the experiments performed}
 \label{tbl:Results}
 \end{center}
 \end{table}
 
 As the table shows, 
%  As we can see, 
 the best results were obtained by Kaldi with a 
 neural network setup. It is interesting to note that the 
 difference between Sphinx and Kaldi is at almost 10\%. This is consistent
 with similar experiments done on Spanish \cite{mena2017automatic}.

%%%%%%%%%%%%%%%%%%%%%%%%%%%%%%%%%%%%%%%%%%%%%%%%%%%%%%%%%%%%%%%%%%%%%%%%%%
\section{Conclusions and future work}\label{sec:conclusion}

This paper described the design of MASRI-HEADSET (MHC), a new corpus 
for Maltese speech aimed to support research on automatic speech recognition (ASR). This 
is the first corpus of its kind for the Maltese language, for which speech
resources are still somewhat limited. 

Preliminary experiments with the 
corpus, while somewhat small-scale, suggest that high-quality acoustic models can be created in spite of 
the small size of the MHC, due to the low noise in the recordings. Our best 
results (WER of $10.56$\%), obtained with a neural network setup, point to the importance of performing future experiments with modern 
speech recognizers based on neural networks such as 
DeepSpeech \cite{amodei2016deep}, Wav2Letter \cite{collobert2016wav2letter}
or Espresso \cite{wang2019espresso}.

The corpus will be freely available from our 
website\footnote{\url{https://www.um.edu.mt/projects/masri/index.html}} for 
research purposes, with additional licensing options for commercial use. 

While MHC is an important first step towards achieving adequate support for Maltese speech technology, it is limited in size. Our current 
work is addressing this issue, as well as other challenges in developing viable ASR technology for spoken Maltese, from multiple angles. 

First, we are studying recent techniques for enhancing ASR in weakly supervised or unsupervised settings, 
with a view to maximising the use of the available data, either via 
pre-training \cite{Schneider2019}, or via augmentation techniques 
such as spectrogram perturbation \cite{Park2019} and noise 
superposition \cite{Hannun2014}. 

Second, a crowdsourcing effort is currently underway, based on the 
Common Voice initiative. Common Voice\footnote{\url{https://voice.mozilla.org/en}} \cite{Roter2019} is an 
open-source, web-based platform created by Mozilla in order to 
crowd-source speech data for various languages. Language communities 
are encouraged to localise the Common Voice website to their 
language, as well as provide a minimum of 5,000 validated sentences 
to be used as prompts to be read out and validated by users. This process is currently 
underway for Maltese, with the website having been completely localised 
and sentence prompts already sampled from the Korpus Malti v3.0. The prompts 
are currently being validated for use. Once this process is complete, 
users will be able to contribute to the construction of a larger, 
community-driven dataset for Maltese by donating their voices to the 
project, as well as validating other users' voice clips. As part of this effort, we are also developing unsupervised techniques designed to 
automatically validate user-contributed data, using heuristics to automatically rank text-speech pairs according to the likelihood that a recorded voice clip matches a textual prompt.

Finally, many of the important challenges for open-domain ASR for Maltese are connected to the complexities arising from the bilingual (Maltese/English) situation described in Section \ref{sec:intro}, where a majority first language with a relatively small number of speakers (in this case Maltese) exists alongside a far better resourced and more widely-spoken language (that is, English). A well-known issue in such intensive language contact situations is the widespread incidence of code-switching and lexical borrowing \cite{Matras2009}. Any viable ASR system for Maltese will also need to handle Maltese-English code-switching. We aim to address this both from the data-collection perspective, through selection of data from naturalistic sources, as well as crowd-sourcing and further data-collection efforts, and from the modelling perspective as we explore transfer learning and learning from mixed language resources.

%%%%%%%%%%%%%%%%%%%%%%%%%%%%%%%%%%%%%%%%%%%%%%%%%%%%%%%%%%%%%%%%%%%%%%%%%%
\section*{Acknowledgements}
Special thanks to Ayrton Brincat for helping us with the final 
revision of the release transcriptions of the corpus.
We also thank KPMG Microsoft Business Solutions (formerly CrimsonWing) 
for providing their grapheme-to-phoneme conversion rules used as the basis for the G2P tool created in this project. The MASRI Project is funded by the University of Malta Research Fund 
Awards. We gratefully acknowledge the support of NVIDIA Corporation with the donation of the Titan Xp used for MASRI research.

%%%%%%%%%%%%%%%%%%%%%%%%%%%%%%%%%%%%%%%%%%%%%%%%%%%%%%%%%%%%%%%%%%%%%%%%%%
% \nocite{*}
\section{References}
\label{main:ref}

\bibliographystyle{lrec}
\bibliography{lrec2020-masri}

\begin{thebibliography}{}

\bibitem[\protect\citename{Amodei \bgroup et al.\egroup }2016]{amodei2016deep}
Amodei, D., Ananthanarayanan, S., Anubhai, R., Bai, J., Battenberg, E., Case,
  C., Casper, J., Catanzaro, B., Cheng, Q., Chen, G., et~al.
\newblock (2016).
\newblock Deep speech 2: End-to-end speech recognition in english and mandarin.
\newblock In {\em International conference on machine learning}, pages
  173--182.

\bibitem[\protect\citename{Azzopardi-Alexander and
  Borg}2013]{azzopardi2013maltese}
Azzopardi-Alexander, M. and Borg, A.
\newblock (2013).
\newblock {\em Maltese}.
\newblock Routledge.

\bibitem[\protect\citename{Besacier \bgroup et al.\egroup
  }2014]{besacier2014automatic}
Besacier, L., Barnard, E., Karpov, A., and Schultz, T.
\newblock (2014).
\newblock Automatic speech recognition for under-resourced languages: A survey.
\newblock {\em Speech Communication}, 56:85--100.

\bibitem[\protect\citename{Borg and Gatt}2017]{borg2017morphological}
Borg, C. and Gatt, A.
\newblock (2017).
\newblock {Morphological analysis for the Maltese language: The challenges of a
  hybrid system.}
\newblock In {\em Proceedings of the 3rd Arabic Natural Language Processing
  Workshop (WANLP'17)}, Valencia, Spain. Association for Computational
  Linguistics.

\bibitem[\protect\citename{Borg \bgroup et al.\egroup
  }2011]{borg2011preparation}
Borg, M., Bugeja, K., Vella, C., Mangion, G., and Gaf{\`a}, C.
\newblock (2011).
\newblock {Preparation of a free-running text corpus for Maltese concatenative
  speech synthesis}.
\newblock In {\em Proceedings of the 3rd International Conference on Maltese
  Linguistics (GhiLM'11)}, Valletta. GhiLM.

\bibitem[\protect\citename{Brincat}2011]{brincat2011}
Brincat, J.
\newblock (2011).
\newblock {\em {Maltese and other languages: A linguistic history of Malta}}.
\newblock Midsea Books, Malta.

\bibitem[\protect\citename{Camilleri}2013]{camilleri2013}
Camilleri, J.~J.
\newblock (2013).
\newblock {A Computational Grammar and Lexicon for Maltese}.
\newblock Master's thesis, Chalmers University of Technology, Gothenburg,
  Sweden.

\bibitem[\protect\citename{{\v{C}}\'{e}pl\"{o}}2018]{bulbul2018}
{\v{C}}\'{e}pl\"{o}, S.
\newblock (2018).
\newblock {\em Constituent order in {Maltese}: A quantitative analysis}.
\newblock {Ph.D.} thesis, Charles University, Prague, Czech Republic.

\bibitem[\protect\citename{Collobert \bgroup et al.\egroup
  }2016]{collobert2016wav2letter}
Collobert, R., Puhrsch, C., and Synnaeve, G.
\newblock (2016).
\newblock Wav2letter: an end-to-end convnet-based speech recognition system.
\newblock {\em arXiv preprint arXiv:1609.03193}.

\bibitem[\protect\citename{Comrie}2009]{comrie2009maltese}
Comrie, B.
\newblock (2009).
\newblock Maltese and the world atlas of language structures.
\newblock {\em Introducing Maltese Linguistics}, pages 3--12.

\bibitem[\protect\citename{Draxler and J{\"a}nsch}2004]{draxler2007speech}
Draxler, C. and J{\"a}nsch, K.
\newblock (2004).
\newblock Speech recorder (version 2.2. 1).
\newblock https://www.bas.uni-muenchen.de/Bas/software/speechrecorder/.

\bibitem[\protect\citename{{European Parliament}}2018]{ep2018}
{European Parliament}.
\newblock (2018).
\newblock Language equality in the digital age ({E}uropean {P}arliament
  {R}esolution 2018/2028({INI})).
\newblock http://www.europarl.europa.eu/doceo/document/TA-8-2018-0332\_EN.pdf.

\bibitem[\protect\citename{Fabri}1993]{Fabri1993}
Fabri, R.
\newblock (1993).
\newblock {\em Kongruenz und die Grammatik des Maltesischen}.
\newblock Niemeyer, Tuebingen.

\bibitem[\protect\citename{Gatt and {\v{C}}{\'e}pl{\"o}}2013]{gatt2013digital}
Gatt, A. and {\v{C}}{\'e}pl{\"o}, S.
\newblock (2013).
\newblock {Digital corpora and other electronic resources for Maltese}.
\newblock {\em Corpus Linguistics 2013}, page~96.

\bibitem[\protect\citename{Gatt and Fabri}2018]{gatt2018productivity}
Gatt, A. and Fabri, R.
\newblock (2018).
\newblock Borrowed affixes and morphological productivity: {A} case study of
  two {M}altese nominalisations.
\newblock In P~Paggio et~al., editors, {\em Languages of Malta: {R}ecent
  Studies}, pages 143--170. Language Science Press, Berlin.

\bibitem[\protect\citename{Grech}2015]{grech2014}
Grech, S.
\newblock (2015).
\newblock {\em Variation in {E}nglish: {P}erception and patterns in the
  identification of {M}altese {E}nglish}.
\newblock {Ph.D.} thesis, University of Malta, Malta.

\bibitem[\protect\citename{Hannun \bgroup et al.\egroup }2014]{Hannun2014}
Hannun, A., Case, C., Casper, J., Catanzaro, B., Diamos, G., Elsen, E.,
  Prenger, R., Satheesh, S., Sengupta, S., Coates, A., and Ng, A.~Y.
\newblock (2014).
\newblock {Deep Speech: Scaling up end-to-end speech recognition}.
\newblock {\em arXiv}, 1412.5567.

\bibitem[\protect\citename{Hoberman}2007]{hoberman2007maltese}
Hoberman, R.~D.
\newblock (2007).
\newblock Maltese morphology.
\newblock {\em Morphologies of Asia and Africa}, 1:257--281.

\bibitem[\protect\citename{Huggins-Daines \bgroup et al.\egroup
  }2006]{huggins2006pocketsphinx}
Huggins-Daines, D., Kumar, M., Chan, A., Black, A.~W., Ravishankar, M., and
  Rudnicky, A.~I.
\newblock (2006).
\newblock {P}ocketsphinx: A free, real-time continuous speech recognition
  system for hand-held devices.
\newblock In {\em 2006 IEEE International Conference on Acoustics Speech and
  Signal Processing Proceedings}, volume~1, pages I--I. IEEE.

\bibitem[\protect\citename{Krauwer}2003]{krauwer2003basic}
Krauwer, S.
\newblock (2003).
\newblock The basic language resource kit (blark) as the first milestone for
  the language resources roadmap.
\newblock {\em Proceedings of SPECOM 2003}, pages 8--15.

\bibitem[\protect\citename{Lamere \bgroup et al.\egroup }2003]{lamere2003cmu}
Lamere, P., Kwok, P., Gouvea, E., Raj, B., Singh, R., Walker, W., Warmuth, M.,
  and Wolf, P.
\newblock (2003).
\newblock The {CMU SPHINX-4} speech recognition system.
\newblock In {\em IEEE Intl. Conf. on Acoustics, Speech and Signal Processing
  (ICASSP 2003), Hong Kong}, volume~1, pages 2--5.

\bibitem[\protect\citename{Lim and Kim}2019]{lim2019integration}
Lim, M. and Kim, J.-H.
\newblock (2019).
\newblock Integration of tensorflow based acoustic model with kaldi wfst
  decoder.
\newblock {\em arXiv preprint arXiv:1906.11018}.

\bibitem[\protect\citename{Matras}2009]{Matras2009}
Matras, Y.
\newblock (2009).
\newblock {\em Language Contact}.
\newblock Cambridge University Press, Cambridge, UK.

\bibitem[\protect\citename{Mena \bgroup et al.\egroup }2017]{mena2017automatic}
Mena, C. D.~H., Ruiz, I. V.~M., and Camacho, J. A.~H.
\newblock (2017).
\newblock Automatic speech recognizers for mexican spanish and its open
  resources.
\newblock {\em Journal of Applied Research and Technology}, 15(3).

\bibitem[\protect\citename{Mena}2019]{mena2019opencor}
Mena, C.
\newblock (2019).
\newblock The ciempiess proper-names pronouncing dictionary.
\newblock In {\em Corpus presented at OpenCor 2019 Conference, Guanajuato City,
  Mexico.[Available online at https://opencor.gitlab.io/corpora-list/]}.

\bibitem[\protect\citename{Micallef}1997]{micallef1997text}
Micallef, P.
\newblock (1997).
\newblock {\em {A text to speech synthesis system for Maltese}}.
\newblock {Ph.D.} thesis, University of Surrey.

\bibitem[\protect\citename{Mifsud}1995]{mifsud1995}
Mifsud, M.
\newblock (1995).
\newblock {\em {Loan verbs in Maltese: A descriptive and comparative study}}.
\newblock Brill, Leiden.

\bibitem[\protect\citename{NSO}2019]{nso2019}
NSO.
\newblock (2019).
\newblock {\em Regional Statistics (Malta), 2019 Edition}.
\newblock National Statistics Office, Malta.

\bibitem[\protect\citename{Panayotov \bgroup et al.\egroup
  }2015]{panayotov2015librispeech}
Panayotov, V., Chen, G., Povey, D., and Khudanpur, S.
\newblock (2015).
\newblock Librispeech: an asr corpus based on public domain audio books.
\newblock In {\em 2015 IEEE International Conference on Acoustics, Speech and
  Signal Processing (ICASSP)}, pages 5206--5210. IEEE.

\bibitem[\protect\citename{Park \bgroup et al.\egroup }2019]{Park2019}
Park, D.~S., Chan, W., Zhang, Y., Chiu, C.-C., Zoph, B., Cubuk, E.~D., and Le,
  Q.~V.
\newblock (2019).
\newblock {SpecAugment: A Simple Data Augmentation Method for Automatic Speech
  Recognition}.
\newblock {\em arXiv}.

\bibitem[\protect\citename{Povey \bgroup et al.\egroup }2011]{povey2011kaldi}
Povey, D., Ghoshal, A., Boulianne, G., Burget, L., Glembek, O., Goel, N.,
  Hannemann, M., Motlicek, P., Qian, Y., Schwarz, P., et~al.
\newblock (2011).
\newblock The {K}aldi speech recognition toolkit.
\newblock In {\em IEEE 2011 workshop on automatic speech recognition and
  understanding}. IEEE Signal Processing Society.

\bibitem[\protect\citename{Ravishankar \bgroup et al.\egroup }2017]{acling2017}
Ravishankar, V., Tyers, F., and Gatt, A.
\newblock (2017).
\newblock {A Morphological analyser for Maltese.}
\newblock {\em Procedia Computer Science (Special Issue on Arabic Computational
  Linguistics)}, 117:175--182.

\bibitem[\protect\citename{Rosner and Joachimsen}2012]{Rosner2012}
Rosner, M. and Joachimsen, J.
\newblock (2012).
\newblock {\em {The Maltese Language in the Digital Age}}.
\newblock Springer, Berlin and Heidelberg.

\bibitem[\protect\citename{Roter}2019]{Roter2019}
Roter, G.
\newblock (2019).
\newblock {Sharing our Common Voices -- Mozilla releases the largest to-date
  public domain transcribed voice dataset}.
\newblock
  https://blog.mozilla.org/blog/2019/02/28/sharing-our-common-voices-mozilla-releases-the-largest-to-date-public-domain-transcribed-voice-dataset/,
  February.

\bibitem[\protect\citename{Schneider \bgroup et al.\egroup
  }2019]{Schneider2019}
Schneider, S., Baevski, A., Collobert, R., and Auli, M.
\newblock (2019).
\newblock {wav2vec: Unsupervised Pre-training for Speech Recognition}.
\newblock In {\em Proceedings of Interspeech 2019}, pages 1--9, Graz, Austria.

\bibitem[\protect\citename{Tiedemann and van~der Plas}2016]{tiedemann2016}
Tiedemann, J. and van~der Plas, L.
\newblock (2016).
\newblock {Bootstrapping a Dependency Parser for Maltese - A Real-World Test
  Case}.
\newblock In Martijn Weiling, et~al., editors, {\em From Semantics to
  Dialectometry : Festschrift in honor of John Nerbonne}. College Publications,
  London.

\bibitem[\protect\citename{Vella}1994]{vella1994}
Vella, A.
\newblock (1994).
\newblock {\em {Prosodic structure and intonation in Maltese and its influence
  on Maltese English}}.
\newblock {Ph.D.} thesis, University of Edinburgh, Scotland.

\bibitem[\protect\citename{Wang \bgroup et al.\egroup }2019]{wang2019espresso}
Wang, Y., Chen, T., Xu, H., Ding, S., Lv, H., Shao, Y., Peng, N., Xie, L.,
  Watanabe, S., and Khudanpur, S.
\newblock (2019).
\newblock Espresso: A fast end-to-end neural speech recognition toolkit.
\newblock {\em arXiv preprint arXiv:1909.08723}.

\bibitem[\protect\citename{Weide}1998]{weide1998cmu}
Weide, R.~L.
\newblock (1998).
\newblock The {CMU} pronouncing dictionary.
\newblock http://svn.code.sf.net/p/cmusphinx/code/trunk/cmudict/.

\bibitem[\protect\citename{Young and Young}1993]{young1993htk}
Young, S.~J. and Young, S.
\newblock (1993).
\newblock {\em The {HTK} hidden Markov model toolkit: Design and philosophy}.
\newblock University of Cambridge, Department of Engineering Cambridge,
  England.

\bibitem[\protect\citename{Zammit}2018]{zammit2018}
Zammit, A.
\newblock (2018).
\newblock A dependency parser for the {M}altese language using deep neural
  networks.
\newblock Master's thesis, Department of Artificial Intelligence, University of
  Malta, Msida, Malta.

\end{thebibliography}

%\section{Language Resource References}
%\label{lr:ref}
%\bibliographystylelanguageresource{lrec}
%\bibliographylanguageresource{lrec2020W-xample}

%%%%%%%%%%%%%%%%%%%%%%%%%%%%%%%%%%%%%%%%%%%%%%%%%%%%%%%%%%%%%%%%%%%%%%%%%%
%\onecolumn 
%\section*{Appendix A: Shared Resources}

 %\begin{table}[ht]
 %\begin{center}
 %\begin{tabular}{|l|}
 %  \hline
 %       {\bf The MASRI-HEADSET Corpus}    \\
 %       {\small \url{www.corpus.com} }     \\
 %       {\bf G2P Tool for Maltese}         \\
 %       {\small \url{www.g2p.com} }        \\
 %       {\bf Sphinx Configuration files}   \\
 %       {\small \url{www.sphinx.com} }     \\
 %       {\bf Kaldi Configuration files}    \\
 %       {\small \url{www.kaldi.com} }      \\
 %   \hline
 %\end{tabular}
 %\caption{Resources shared in this paper}
 %\label{tbl:SharedResources}
 %\end{center}
 %\end{table}

%%%%%%%%%%%%%%%%%%%%%%%%%%%%%%%%%%%%%%%%%%%%%%%%%%%%%%%%%%%%%%%%%%%%%%%%%%
\onecolumn 
\section*{Appendix A: The Maltese Phonology}

 \begin{table*}[ht]
 \begin{center}
 \begin{tabular}{|l|c|c|c|c|c|c|c|}
   \hline
   Consonants           &Bilabial&Labiodental&Alveolar    &Postalveolar           & Palatal              &Velar                  & Glottal              \\
   \hline
   Plosive (voiceless)  &   p    &           &    t       &                       &                      &  k                    &\textglotstop         \\
   Plosive (voiced)     &   b    &           &    d       &                       &                      &  g                    &                      \\
   Affricate (voiceless)&        &           &\texttslig  &                       &\textteshlig          &                       &                      \\
   Affricate (voiced)   &        &           &\textdzlig  &                       &\textdyoghlig         &                       &                      \\
   Fricative (voiceless)&        &    f      &    s       &\textesh               &                      &                       &    h                 \\
   Fricative (voiced)   &        &    v      &    z       &\textyogh              &                      &                       &                      \\
   Nasal (voiced)       &   m    &           &    n       &                       &                      &                       &                      \\
   Lateral Approximant (voiced)& &           &    l       &                       &                      &                       &                      \\
   Trill (voiced)       &        &           &    r       &                       &                      &                       &                      \\
   Approximant (voiced) &        &           &            &                       &     j                &  w                    &                      \\
   \hline
   Vowels               &        &           & Front      &                       &   Central            &                       &    Back              \\
   \hline
   Close                &        &           &   i        &                       &                      &                       &                      \\
                        &        &           &            & \textsci              &                      &\textupsilon           &                      \\
   Close-mid            &        &           &            &                       &                      &                       &                      \\
                        &        &           &            &                       &                      &                       &                      \\
   Open-mid             &        &           &            &\textepsilon           &                      &                       & \textopeno           \\
                        &        &           &            &                       & \textturna           &                       &                      \\
   Open                 &        &           &            &                       &                      &                       &                      \\
   \hline
   Long Vowels          &        &           & Front      &                       &   Central            &                       &    Back              \\
   \hline
   Close                &        &           &i\textipa{:}&                       &                      &                       &                      \\
                        &        &           &            &\textsci\textipa{:}    &                      &\textupsilon\textipa{:}&                      \\
   Close-mid            &        &           &            &                       &                      &                       &                      \\
                        &        &           &            &                       &                      &                       &                      \\
   Open-mid             &        &           &            &\textepsilon\textipa{:}&                      &                       & \textopeno\textipa{:}\\
                        &        &           &            &                       & \textturna\textipa{:}&                       &                      \\
   Open                 &        &           &            &                       &                      &                       &                      \\
   
   \hline
   {\small Low-Pitch Accented Vowels} &&     & Front      &                       &   Central            &                       &    Back              \\
   \hline
   Close                &        &           &    \`i     &                       &                      &                       &                      \\
                        &        &           &            &                       &                      &       \`u             &                      \\
   Close-mid            &        &           &            &                       &                      &                       &                      \\
                        &        &           &            &                       &                      &                       &                      \\
   Open-mid             &        &           &            &   \`e                 &                      &                       &      \`o             \\
                        &        &           &            &                       &       \`a            &                       &                      \\
   Open                 &        &           &            &                       &                      &                       &                      \\
   \hline
 \end{tabular}
 \caption{The Phonological System of the Maltese Language}
 \label{tbl:Maltese_Phonology_Table}
 \end{center}
 \end{table*}

\end{document}